# Gesture Recognition from Skeleton Data for Intuitive Human-Machine Interaction

André BRÁS[1], Miguel SIMÃO and Pedro NETO

*Centre for Mechanical Engineering, Materials and Processes, University of Coimbra*

**Abstract.** Human gesture recognition has assumed a capital role in industrial applications, such as Human-Machine Interaction. We propose an approach for segmentation and classification of dynamic gestures based on a set of handcrafted features, which are drawn from the skeleton data provided by the Kinect sensor. The module for gesture detection relies on a feedforward neural network which performs framewise binary classification. The method for gesture recognition applies a sliding window, which extracts information from both the spatial and temporal dimensions. Then we combine windows of varying durations to get a multi-temporal scale approach and an additional gain in performance. Encouraged by the recent success of Recurrent Neural Networks for time series domains, we also propose a method for simultaneous gesture segmentation and classification based on the bidirectional Long Short-Term Memory cells, which have shown ability for learning the temporal relationships on long temporal scales. We evaluate all the different approaches on the dataset published for the ChaLearn Looking at People Challenge 2014. The most effective method achieves a Jaccard index of 0.75, which suggests a performance almost on pair with that presented by the state-of-the-art techniques. At the end, the recognized gestures are used to interact with a collaborative robot.

## Introduction

Recently, human action and gesture recognition has attracted increasing attention of researchers and has played a significant role in areas such as Human-Machine Interaction (HMI). Intuitive and reliable communication between humans and robots is essential for a successful collaboration. Regarding natural interfaces, the most relevant channels for communication between humans and robotic assistants are voice and gestures [1]. Since the typically noisy industrial environments render verbal interaction ineffective, gestures have been the most explored way of communication to collaborate with robots [2].

Although important advances have been made in sensor technologies and machine learning methods, automated gesture segmentation and classification remain a challenging problem [3]. The problem of recognizing gestures comprises many difficulties, including noisy and missing data, variability across individuals, irregular observation conditions (e.g. lighting, background and viewpoint) and infinite out-of-vocabulary motions. Moreover, the understanding of video content for gesture recognition persists a growing area of research due to the higher complexity coming with the temporal dimension [4]–[6].

---

[1] andre.bras@uc.pt

The appearance of the Kinect sensor was a remarkable step for computer vision and it is applied to a variety of tasks involving gesture recognition, namely entertainment and HMI. As the Kinect provides built-in skeleton data along with high-resolution depth and color images, new datasets emerged, providing researchers with the chance to design novel methods and test them on a larger number of sequences. We present experiments on the dataset published for the ChaLearn Looking at People (LAP) Challenge 2014 [7].

This paper focuses on labelling of video sequences. Our approach relies on a set of handcrafted features drawn from the skeleton data. We propose an effective module for gesture localization, which performs framewise binary classification. The method for gesture recognition applies a scheme based on a sliding window. We also combine different windows to get a multi-temporal scale approach. Furthermore, we present a method for simultaneous gesture detection and classification, which uses a Recurrent Neural Network (RNN) with bidirectional Long Short-Term Memory (LSTM) cells [8]. These recurrent cells ease the learning of temporal relationships on long temporal scales and have shown slightly better performance than standard cells [9].

**1. Related work**

Traditional approaches for gesture recognition typically include spatio-temporal engineered descriptors followed by classification. Even the most accurate methods submitted for the ChaLearn LAP Challenge 2014 proposed handcrafted descriptors for the extraction of features. The method with the best score learns features from each visual modality, but those built from the skeleton data are manually extracted features [4]. The second most precise method relies entirely on handcrafted features and the global appearance of each gesture is inferred by those drawn from the skeleton data [5].

While many human gestures may be distinguished by the position and movements of main joints, such as the elbows and shoulders, other gestures differ primarily in hand pose and its positioning relative to the body or face. Consequently, the Histogram of Oriented Gradients (HOG) [10] is a handcrafted feature descriptor often applied to discriminate the hand pose with notable results in gesture recognition [5].

For decades, constructing a machine learning system required careful engineering and considerable domain expertise to design a feature extractor. Furthermore, the choice of features is a difficult task since they are highly problem-dependent. Deep learning methods are representation learning methods, allowing then a machine to be fed with raw data and automatically discover the representations needed for detection or classification tasks [11]. The learned features have shown valuable results in a myriad of domains, being many times better than those achieved with engineered descriptors [12].

Convolutional Neural Networks (CNNs) [13] are examples of deep networks which can be used to learn latent and complex features directly from data. The use of CNNs to almost halve the error rate for object recognition was a breakthrough that precipitated the adoption of deep learning by the computer vision community [14]. Di Wu *et al.* used deep networks, including a 3D CNN to handle color and depth images, to model the emission probabilities for a Hidden Markov Model (HMM) [6]. Even though CNNs have been applied for feature construction, they can be trained end to end from pixel values to classification outputs. Ji *et al.* applied these networks for human action recognition in airport surveillance videos [15]. Karpathy *et al.* also studied multiple approaches for extending the connectivity of a CNN in time domain [16]. Their *Slow Fusion* model, which achieved the best performance on the Sports-1M dataset, is again a 3D CNN.

Pigou *et al.* showed that the temporal pooling is not sufficient for gesture recognition, where temporal information is more discriminative than in general video classification tasks [9]. They used spatio-temporal convolutions followed by bidirectional recurrence (RNN with LSTM cells) and they achieved state-of-the-art results on the ChaLearn LAP Challenge 2014. The concept of combining the exclusive feature extraction capability of CNNs with the modeling of temporal dynamics by LSTM networks also provided state-of-the-art results for human activity recognition using multi-modal wearable sensors [17].

**2. Gesture segmentation**

*2.1. Pose descriptor*

Central to our approach is a pose descriptor of handcrafted features, which are based on the skeleton data provided by the Kinect sensor for the 11 human body joints depicted in Figure 1.1. This pose descriptor comprises spatial information and temporal details, such as speed and acceleration, of a short time window around the current frame.

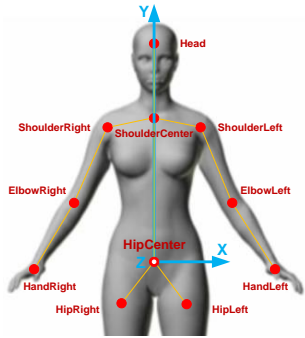
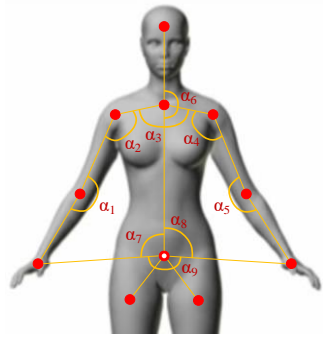

**Figure 1.1.** Relevant body joints.   **Figure 1.2.** Angles formed by the linked joints.

Our method for gesture detection initially follows the procedure proposed by Zanfir *et al.* [18]. Accordingly, we compute a normalized pose vector, as well as the velocities and accelerations for all the 11 joints. Figure 1.1 illustrates the upper body joints as a tree structure, where the HipCenter joint is a root node. Its raw coordinates are subtracted from all the other position vectors to decrease the influence assumed by the spatial position of the body. The normalization of the distance between each pair of joints is also applied to balance the differences in proportions of the users' bodies. Then, each coordinate is smoothed along the temporal dimension with a 5 by 1 Gaussian filter which uses a standard deviation equal to 1. Finally, the velocities and accelerations for each joint are the first and second order derivatives of the correspondent smoothed positions.

We achieve a more precise descriptor collecting characteristic angles and pairwise distances [4]. Inclination angles are formed by all triples of anatomically connected joints if we add two virtual bones: RightHand/LeftHand – HipCenter (Figure 1.2). Azimuth angles provide information about the pose appearance in a coordinate system associated with the body and they are computed to the same triples of joints used by the inclination angles. Bending angles are measured between a vector perpendicular to the torso and the normalized position of each joint. The 55 distances between each pair of joints are the last feature added to the descriptor. Combining all features, which are normalized to zero mean and unit variance, we get a 183-dimensional pose descriptor for each frame.

*2.2. Supervised segmentation*

Different gestures can be very similar in their initial or final stages and framewise classification is often noisy or even erroneous. We introduce this module to prevent those negative effects. The classifier distinguishes resting moments from periods of activity and it can spot starting and ending points of each gesture. This stage is implemented based on the handcrafted descriptor. All training frames labeled with a gesture class are used as positive examples and a set of frames before and after such gesture as negatives.

The architecture chosen to accomplish our goal is presented in Figure 2. Since the model is trained with the pose descriptors, its input layer has 183 units. Furthermore, the network comprises 2 hidden layers and both contain 100 units. However, while the first one applies the Rectified Linear Unit (ReLU) as activation function, the second one employs the hyperbolic tangent. The output layer has a single neuron and uses the sigmoid activation function. This layer outputs 1 when dealing with periods of activity and 0 otherwise. The network is optimized with the Scaled Conjugate Gradient (SCG).

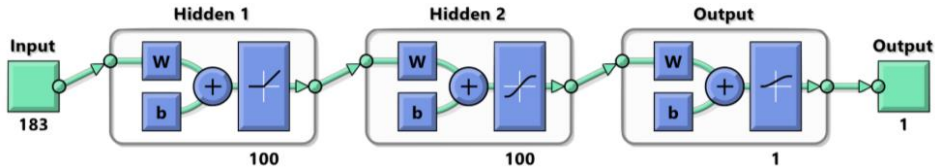

**Figure 2.** Network architecture for the segmentation module. The input layer uses 183 features, the hidden layers have 100 units each and the output layer owns a single unit.

For each sample, the neural network outputs as many scores as the number of frames on the sample. The predictions are unstable and, therefore, we smooth them through local regression with the least squares method and a second-degree polynomial. After that, we use a threshold to establish that all frames with a score higher than 0.4 belong to a period of activity. Furthermore, we only consider periods of activity that span at least 12 frames.

**3. Gesture classification**

Framewise classification is a reasonable strategy for researchers coping with static gestures, as the data of a single frame are very particular. However, our work points to dynamic gestures and we develop 3 appropriate models. Methods A and B are approaches based on a sliding window which can concatenate spatial information from different moments. These models focus only on the periods of activity segmented by the model in the section above. Method C is a deep model testing a RNN with LSTM cells.

*3.1. Method A*

The application of a window leads to a dynamic pose, which is a sequence of descriptors sampled at a given temporal step and concatenated to produce a spatio-temporal vector. Our Method A builds dynamic poses with 3 pose descriptors, being the step between the selected frames equal to 4. Hence, this window embraces exactly 9 frames. The step of the window after each application is equal to 2. Furthermore, we want to collect at least 5 dynamic poses from each period of activity, ensuring thus the classifier has enough data to output a robust prediction. When the dimension of a period is insufficient to meet that condition, the data is resized to a minimum length using cubic interpolation.

The dynamic poses built according to the description above are taken as input by a feedforward network, whose architecture is shown in Figure 3. Each dynamic pose includes 3 pose descriptors and, therefore, the input layer has 549 units. The network comprises 2 hidden layers, being the first one composed by 300 units and the second one by only 100 units. Both layers apply the hyperbolic tangent as activation function. The output layer has 20 units, which is the number of classes, and uses the softmax activation function. This network is also trained with the SCG.

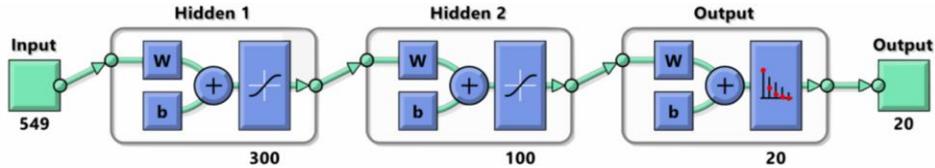

**Figure 3.** Network architecture for the classification module. The input layer uses 549 features, the first and second hidden layers have 300 and 100 units, respectively, and the output layer owns 20 units.

The final details of the classification process can be different between periods of activity depending on its length. According to the study of the mean length of gestures, we assume that a period of movement with less than 55 frames contains just a single gesture. In this case, we slide the window over the whole segment and the network evaluates the dynamic poses individually. For each dynamic pose, if the largest score is higher than a threshold, which is equal to 0.8717, we record the corresponding class. At the end, if the most common class gets the absolute majority between the recorded classes, we classify the period with that class. Otherwise, the segment remains unlabeled. When the period has a larger length, we assume that it includes multiple gestures and the goal is to detect each one of them. Once again, if the score of the winning class is greater than a threshold, which is now equal to 0.6255, we record that class. Whenever the same gesture is identified in at least 3 consecutive windows, we assume that this gesture was effectively performed. Then, all frames encompassed by the consecutive windows classified with the same class are labeled with this class.

*3.2. Method B*

Too wide dynamic poses lead to noisy predictions, particularly in the initial and final stages of gestures. On the other hand, short dynamic poses are not discriminative, since some gesture classes have similar appearance. Searching for different combinations of hyperparameters improves some predictions but worsens others. Therefore, we introduce a robust method that applies 3 sliding windows, which are very similar to that one used by Method A, but create dynamic poses by sampling descriptors with different time steps. This association results in a model that accommodates multiple temporal scales.

According to the paragraph above, this method employs 3 sliding windows. The first one collects pose descriptors with a step between the selected frames equal to 4, the second one uses a step of 3 and the last one applies a step equal to 2. All the other hyperparameters monitoring the application of each sliding window are the same than those defined above. This model involves the training of 3 neural networks and the architecture of each one is equal to that one depicted in Figure 3.

When the windows are slid over each period of activity, we get 3 sets of scores. They are combined with optimized weights: 0.4895, 0.4576 and 0.0529, respectively. Then, the classification process follows the guidelines defined above. However, the threshold used for short periods is updated to 0.6014 and the other is set to 0.6033.

*3.3. Method C*

The third method implemented in this paper is based on a RNN with bidirectional LSTM cells. The procedure is different from those covered above, since we discard all the information from the segmentation section. We slid a window with 10 frames of length over each training sample, collecting data from all frames inside the window. Thereby, for each iteration, we get a matrix whose dimensions are the number of frames by the number of features. If all frames within the window belong to a resting moment, the step between iterations is equal to 5 frames. Otherwise, the step is only of 2 frames.

The deep network here applied comprises 3 hidden layers with bidirectional LSTM cells, 2 dropout layers and 1 dense layer with fully connected units. The first and second hidden layers, which have 1024 cells each and use leaky ReLU activation function, are followed by dropout layers, whose probability to drop out input elements is equal to 60%. The third LSTM layer has 512 units. The dense layer is composed by 21 units, which is the number of classes (including the resting moments), and uses the softmax activation function. This network is optimized using the Stochastic Gradient Descent with Momentum (SGDM). The learning rate is initialized with 0.01 and it is dropped by a factor of 0.85 after each set of 10 epochs until a maximum of 150 epochs. We also use batches of 128 samples of data to accelerate the convergence and prevent overfitting.

At the end, we collect sequences of data from the test samples with a step after each iteration equal to the length of the window, which removes the overlap between sequences. The predictions are again unsteady and they need to be smoothed, as we did for gesture segmentation. Finally, we only consider that a gesture was effectively performed if there are at least 15 consecutive frames classified with the same label.

**4. Experiments**

*4.1. ChaLearn Looking at People Challenge 2014*

In 2014, ChaLearn proposed a competition for multi-modal gesture recognition. The dataset includes nearly 14,000 gestures belonging to a vocabulary of 20 classes selected from Italian sign language. It also contains multiple out-of-vocabulary motions. This dataset provides RGB-D videos, as well as the skeleton joints data. Furthermore, the dataset is split into training, validation and test sets. Even though the ground truth for the validation and test sets has been released, it is only used for the evaluation of each model.

We follow the evaluation procedure proposed by the competition organizers and, therefore, we use the Jaccard index to quantify the performance of the models. Thus, for the $n$-th gesture in sequence $s$, the Jaccard index is defined as:

$$J_{s,n} = \frac{A_{s,n} \cap B_{s,n}}{A_{s,n} \cup B_{s,n}} \qquad (1)$$

where $A_{s,n}$ is the ground truth of gesture $n$ in sequence $s$ and $B_{s,n}$ is the prediction for such gesture in the same sequence. $A_{s,n}$ and $B_{s,n}$ are binary vectors where the frames in which the given gesture is being performed are set to 1. The overall performance is computed as the mean Jaccard index among all categories and all sequences.

*4.2. Gesture segmentation*

We tested several architectures for the segmentation model, namely networks with different number of hidden layers and different number of units in these layers. We realized that networks with 2 hidden layers behave slightly better than networks with a single hidden layer. Furthermore, we noticed that the best performance was achieved using the ReLU as activation function for the first hidden layer and the hyperbolic tangent for the second one. The strategy detailed in Section 2.2 allows the classification of each frame with an accuracy of 96.8%. Figure 4.1 shows an example of the segmentation.

When the users perform gestures separated by a clear time interval, the segmentation model works well. However, some subjects execute consecutive gestures and others adopt a dynamic pose along the whole sample. In these cases, the segmentation is challenging and the model is only able to detect a big period of activity. This is the reason why we described a different classification process in Section 3.1 for larger periods of activity.

As we said before, the dataset includes multiple out-of-vocabulary gestures. However, the segmentation model detects any significant movement of the user because it does not know what gesture is being performed, as it can be seen at the end of Figure 4.2. Thus, the unlabeled gestures are a task for the classification models, which must be accurate enough to predict that those sequences do not belong to the vocabulary.

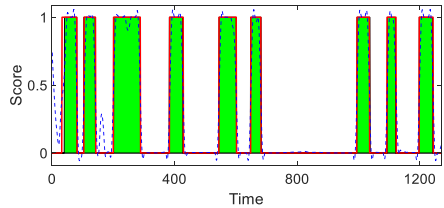 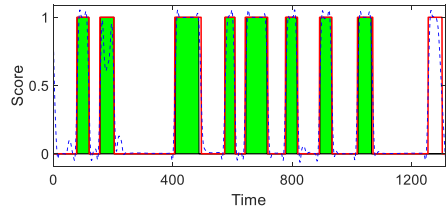

**Figure 4.1.** The model accurately segments each gesture. The shaded area is the ground truth, the blue dashed line is the smoothed score and the red solid line represents the final segmentation.

**Figure 4.2.** Our segmentation model, which is not aware of gesture classes, detects the out-of-vocabulary gestures just as it performs with labeled gestures.

*4.3. Gesture classification and robot control*

In Table 1, we compare our proposed methods to the top 3 performances of the ChaLearn LAP Challenge 2014. All these scores were achieved using only skeleton data.

**Table 1.** Performance of 3 out of the 17 submissions for the ChaLearn LAP Challenge 2014. These scores can be matched with the best Jaccard index achieved by each of our models.

| Method | Score |
| --- | --- |
| [4] | 0.8080 |
| [19] | 0.7948 |
| [5] | 0.7910 |
| A | 0.6928 |
| B | 0.7047 |
| C | 0.7467 |

Our Method A for gesture classification, which builds dynamic poses from a single window, achieves a Jaccard index of 0.6928. The introduction of multiple temporal scales performed in Method B allows an improvement in the score, which increases to 0.7047. We can conclude that the combination of different windows is suited to accommodate the variances in performances presented by different users. Furthermore, the weights used in Method B to obtain a single vector of scores suggest that the discriminative power of dynamic poses depends on the step between the selected frames and it is maximum for broader windows. The deep learning approach summarized in Method C reaches a Jaccard index of 0.7467, which is our top score.

The confusion matrix (in log form) in Figure 5 illustrates the performance of Method C. Some gestures are more easily recognized than others. For instance, the class of gestures #16 is only sporadically confused with class #20. This happens because these classes include identical movements, which do not resemble with the actions performed in the remaining gestures. On the other hand, the class #3 is commonly confused with many gestures. We can also observe that classes #14 and #15 are confounded with each other. Figure 6 shows the similarities that lead to this mislabeling. Lastly, the column on the right represents the false negatives, which are frames labeled with some gesture class but classified as a resting moment. So, there are gestures that are not being detected. On the other hand, the last line denotes the false positives, which are frames belonging to a resting moment but classified with some gesture class. The main reason for this fact are the out-of-vocabulary gestures performed to mislead the model.

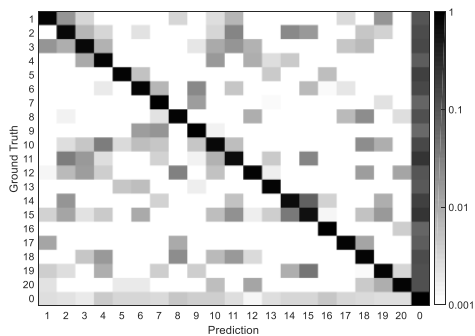

**Figure 5.** Confusion matrix computed with the predictions achieved by Method C.

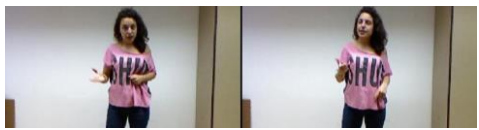

**Figure 6.** Example of gestures that differ primarily in hand pose and, therefore, the skeleton data is insufficient to differentiate them. On the left, the user is performing the gesture #14 and on the right the gesture #15.

We also present the ground truth labels and the predictions for sample #703 in Figure 7. We chose this sample to make possible a direct comparison with the output achieved by Di Wu *et al.* [6]. The top image in Figure 7 shows that this video includes 10 labeled gestures. As we mentioned earlier, it is possible to confirm that there are gestures separated by a clear time interval, but there are also consecutive instances of different gestures. Our model overcomes this issue and correctly predicts all labels, as well as accurately spots starting and ending points of each gesture. On the other hand, the method labels an out-of-vocabulary gesture. Even this behavior represents a good conclusion because the user performs at least 4 out-of-vocabulary gestures in this video. We notice that our model achieves a far better prediction for this sample than that presented by Di Wu *et al.* [6]. Despite this precise prediction, this sample only achieves a Jaccard index of 0.8672, which proves the burdensome nature of this index.

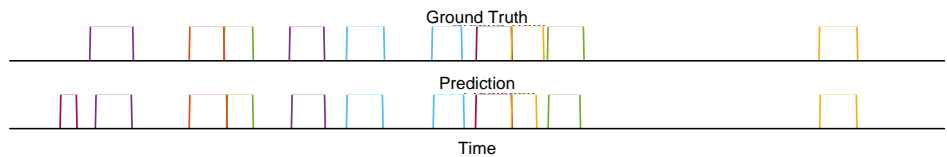

**Figure 7.** Comparison between the ground truth labels and the predictions computed by Method C for a sample of the test set. Classes are differentiated by colors.

We present a Human-Robot Interface (HRI) with a collaborative industrial robot. Our goal is the development of a real process of interaction to complete a useful task. Here, our priority is to capture and move a tool, which includes the following 5 actions: 1) the movement of the robot to the tool surroundings; 2) the closure of the gripper; 3) the translation of the tool; 4) the opening of the gripper; 5) and the return to the default position. For this purpose, we trained a model with only 5 out of 20 gestures included in the dataset used before. A visual description of this task is shown in Figure 8.

This process is natural because only a few intuitive and easy to learn gestures are required. This connection to collaborative robotics represents a breakthrough, since no researcher exploring ChaLearn LAP Challenge 2014 published such type of application. We use KUKA Sunrise Toolbox (KST), which is a MATLAB toolbox, to operate KUKA Sunrise.OS controller [20].

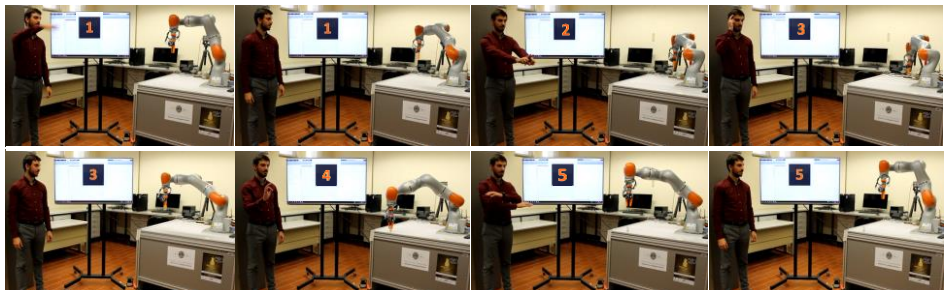

**Figure 8.** Different stages of a HMI, which are arranged from left to right and top to down. The user performs 5 selected and each one leads to a specific action accomplished by the robot. (https://youtu.be/2S-z3WCAAAU)

## 5. Conclusion

In this work, we presented a model for gesture detection, which is based on a descriptor of handcrafted features. It can differentiate resting moments from periods of activity with an accuracy of 96.8%. We also reported 3 different approaches for gesture classification based on the same descriptor. Our Method C applies a RNN with bidirectional LSTM cells to perform simultaneous detection and classification of gestures. This strategy achieved a Jaccard index of 0.7467. Finally, we built a short version of Method C with only 5 out of 20 gestures in the dataset, which we used as commands for a robot.

An important section of computer vision community is aligning its efforts to deep learning approaches. Therefore, future work will be done to replace the handcrafted features with automatically learned features. In particular, we want to apply CNNs to accomplish this task. In addition, it is also our intention to extend the input channels to RGB-D images, making the representations of each gesture more robust.

## Acknowledgement

This research was supported by Portugal 2020 project DM4Manufacturing POCI-01-0145-FEDER-016418 by UE/FEDER through the program COMPETE2020.